\pdfoutput=1

\documentclass[11pt]{article}

\usepackage[preprint]{acl}

\usepackage{times}
\usepackage{latexsym}
\usepackage{enumitem}
\usepackage[T1]{fontenc}

\usepackage[utf8]{inputenc}

\usepackage{microtype}

\usepackage{inconsolata}

\usepackage{graphicx}

\title{ Beyond Agreement: Rethinking Ground Truth in Educational AI Annotation}

\author{Danielle R. Thomas \\
  Carnegie Mellon University \\
  Pittsburgh, PA \\
  \texttt{drthomas@cmu.edu} \\\And
  Conrad Borchers \\
  Carnegie Mellon University \\
  Pittsburgh, PA \\
  \texttt{cborchers@cmu.edu} \\\And Kenneth R. Koedinger \\
  Carnegie Mellon University \\
  Pittsburgh, PA \\
  \texttt{koedinger@cmu.edu}}

\begin{document}
\maketitle
\begin{abstract}
Humans can be notoriously imperfect evaluators. They are often biased, unreliable, and unfit to define "ground truth." Yet, given the surging need to produce large amounts of training data in educational applications using AI, traditional inter-rater reliability (IRR) metrics like Cohen’s kappa remain central to validating labeled data. IRR remains a cornerstone of many machine learning pipelines for educational data. Take, for example, the classification of tutors’ moves in dialogues or labeling open responses in machine-graded assessments. This position paper argues that overreliance on human IRR as a gatekeeper for annotation quality hampers progress in classifying data in ways that are valid and predictive in relation to improving learning. To address this issue, we highlight five examples of complementary evaluation methods, such as multi-label annotation schemes, expert-based approaches, and close-the-loop validity. We argue that these approaches are in a better position to produce training data and subsequent models that produce improved student learning and more actionable insights than IRR approaches alone. We also emphasize the importance of external validity, for example, by establishing a procedure of validating tutor moves and demonstrating that it works across many categories of tutor actions (e.g., providing hints). We call on the field to rethink annotation quality and ground truth—prioritizing validity and educational impact over consensus alone.
\end{abstract}

\section{Introduction}

In the field of educational measurement, researchers increasingly rely on automated assessment methods to enable scalable, real-time evaluation of learning \cite{messer2024automated,thomas2025does}. These methods typically include artificial intelligence (AI)-based models, such as the BLEURT \cite{BleuRT} metric for machine translation, and recent rubric-based prompting strategies with large language models (LLMs) \cite{LLM-Rubric}. Yet, even in such scenarios, the ultimate source of truth remains tied to human annotations, either as a source of supervision during model training or as a post-hoc verification of model outputs. As the quality of any evaluation protocol hinges on the robustness of its underlying “ground truth," human annotation schemes must not be taken at face value, but instead be rigorously assessed for pertinence and alignment with educational evaluation objectives.

This work focuses on the limitations of inter-rater reliability (IRR) and recommends alternatives. IRR is a statistical measure used to determine the degree of agreement among multiple human annotators \cite{gwet2021handbook} and is often used as a basis for validation in educational evaluation and modeling. Although IRR provides a convenient quantitative proxy for annotation quality, it is fundamentally limited by the subjective, biased, and often inconsistent nature of human judgment \cite{gwet2021handbook}. 

The concerns surrounding IRR in the field of educational AI are not new. Researchers have long challenged the sufficiency of traditional IRR metrics such as Cohen’s kappa and Krippendorff’s $\alpha$ in capturing the complexity of human evaluative tasks \cite{Ando2005,gwet2021handbook,doewes2023evaluating,plank2022problem}. The data labeling industry echoes this concern, “There is a general belief in the data labeling industry that high inter-rater reliability indicates high-quality data... However, this is not always the case” \cite{toloka2023irr}. Yet, educational applications are also different from this prior research: As we go on to argue, the impact of assessment on learning and its generalizability across learning contexts poses distinct opportunities to move beyond IRR-only evaluations. The present study's primary contribution is to provide a comprehensive understanding of the problem of IRR-based validation and the paths to move beyond it in educational applications.

\subsection{Understanding the Problem}

In education and beyond, the reliability of human-annotated training data as the ultimate "gold standard" for evaluating learners’ responses has come under increasing scrutiny \cite{chen2024humans,messer2024consistent}. While these issues—such as inconsistency, subjective interpretation, and bias—are well documented, we argue they are increasingly overlooked in the rush to use capable AI models to scale assessments quickly. IRR offers a tempting sense of rigor, often mistaken for objectivity, especially in inherently subjective tasks like grading essays or labeling open responses. However, high IRR can mask annotation shallowness, promote premature consensus, and ignore valid alternative interpretations. Overreliance on IRR is common and problematic, reinforced by decades of practice but increasingly out of step with the complexity of contemporary assessment tasks and AI capabilities \cite{Ando2005,doewes2023evaluating,gwet2021handbook}. The need for more robust, nuanced, and task-sensitive evaluation strategies is now more urgent than ever, especially as new educational AI models are rapidly being deployed.

Next to the intrinsic limitations of AI systems—such as hallucinations in LLMs or the low interpretability of many ML architectures \cite{huang2025survey}—there are deep-seated problems stemming from the training data itself. This includes both the human-annotated datasets and the open-web corpora typically used to train foundation models (e.g., LLMs are known to include biases they soaked up from these training data \cite{chen2024humans}. Our work focuses specifically on the former: when training data for assessment models is generated by humans, IRR is often used to validate annotation rubrics and establish the “gold standard” against which models are evaluated. However, IRR does not always reflect annotation correctness or task difficulty. High agreement can obscure flawed or superficial annotations, while disagreement may indicate productive ambiguity or meaningful variation in interpretation. This paper highlights alternative approaches to supplement and strengthen traditional methods, highlighting a multidimensional approach that can support the demands of scalable, equitable, and informative educational assessment.

\subsection{Aims}
The stakes are high. Presently, 86\% of students and 60\% of teachers report using AI tools, as the education market is expected to exceed \$88 billion by 2032 \cite{digitaleducationcouncil2025survey}. AI has become ubiquitous within classrooms, interventions, and high-stakes testing, with the need to responsibly define and evaluate "ground truth" increasingly urgent. If we continue to equate consensus with correctness, we risk optimizing models not for pedagogical utility or learning outcomes, but for compliance with flawed human standards. We argue that the field of educational AI must move beyond narrow agreement metrics and embrace more flexible, validity-driven approaches to annotation that ensure effectiveness and impact of AI tools.

This work contributes examples of alternative approaches proposed by researchers in the field and their results, such as multi-label annotations that reflect interpretative diversity and close-the-loop validity measures that tie labels to learning outcomes. Together, these approaches provide a richer, more responsible framework for defining and validating ground truth in educational AI. However, these alternative approaches alone are not the solution but provide some examples of how other researchers have attempted to varying degrees of success to overcome the challenges of sole reliance on IRR alone. In showcasing other approaches, we emphasize the lack of external validity in educational AI, such as tutoring systems. We hope to increase awareness as educational AI systems are encroaching on learning as we know it. 

The aims of this work are three-fold:

\begin{itemize}
\item \textbf{Challenge the field's overreliance on inter-rater reliability (IRR)} as the primary validator of annotation quality in educational AI, arguing that consensus alone is insufficient for modeling complex, subjective data 

\item \textbf{Introduce and illustrate alternative or supplemental frameworks} that support a more multidimensional, validity-centered approach to defining “ground truth” in assessment

\item \textbf{Call attention to the lack of external validity in educational AI} and propose a challenge of demonstrating examples in the field, such as a generalizable tutoring model across a diverse range of datasets.
\end{itemize}

\section{Case Applications in Educational AI}
\subsection{Comparative Judgment}
\textit{Reported Use Case: Using comparative judgment to assess students’ reading comprehension and fluency in open responses.} \citet{henkel2023leveraging} present a compelling alternative to traditional IRR approaches by implementing comparative judgment as a method for labeling educational data in the form of middle school students' open response to math problems. The researchers identify the limitations of relying on expert raters and rigid categorical rubrics for scoring student responses, especially for complex or open-ended tasks, and propose comparative judgment as a more scalable, accessible alternative. Comparative judgment requires raters to determine which of two student responses is better, rather than assigning an absolute score. This approach is cognitively easier for raters, particularly non-experts, and aligns with reinforcement learning from human feedback methods used in AI \cite{RLHFOriginalPaper}. In two experiments involving short-answer reading comprehension and oral reading fluency, the study compares traditional categorical judgment with comparative judgment. Results show that comparative judgment substantially improved both accuracy and inter-rater reliability. For short-answer tasks, Krippendorff’s $\alpha$ improved from 0.66 to 0.80, and accuracy increased by 13\%. For oral fluency, $\alpha$ improved from 0.70 to 0.78. These gains were statistically significant.

\citet{henkel2023leveraging} argue that comparative judgment not only improves labeling quality but also challenges the primacy of IRR as the sole measure of annotation quality. They demonstrate that comparative approaches can match or exceed expert-level consistency, even when crowdworkers are used. This study makes a significant contribution by showing that comparative judgment can be an effective alternative or supplement to IRR in educational data annotation, with practical implications for scaling data labeling efforts in educational research and AI development.

\subsection{Multi-label Annotation}
\textit{Reported Use Case: Identifying toxic or offensive text in chat messages.} \citet{arhin2021ground} proposed a multi-label annotation strategy to address the challenges of subjectivity and inconsistency in toxic text classification. Although this use case is not directly applied to an education context, per se, identifying possibly harmful language is important in all aspects of educational AI. Recognizing that language is deeply contextual and that annotator judgments often vary due to differing backgrounds, values, and interpretations, the authors rejected the traditional assumption of a singular, definitive ground truth label. To operationalize their approach, they re-annotated three toxic text datasets using three context-based label types: \textbf{strict label} (based on the presence of offensive words, regardless of context); \textbf{relaxed label} (a more lenient judgment allowing for interpretive variability); \textbf{inferred group label} (based on how a statement might be perceived if uttered by a member of the referenced group). This multi-labeling scheme captured nuanced perspectives and better represented the diversity of valid interpretations. While the approach did not always improve inter-annotator agreement, it produced annotations with higher alignment to external machine-learning classifiers (e.g., Detoxify and Perspective API), thus offering enhanced dataset quality and potential model generalizability. Researchers emphasized that multi-labeling reflects the inherent ambiguity in human language more faithfully than forced consensus and serves as a more robust foundation for training and evaluating AI systems \cite{arhin2021ground}. 

\subsection{Expert-Based Labeling Approaches}
\textit{Reported Use Case: Evaluating annotator quality through expert-grounded benchmarks in dialogues}. While traditional IRR assumes agreement among annotators as the gold standard, recent work challenges this assumption by leveraging expert-labeled data as a more principled benchmark for evaluating annotation quality. We provide two examples of leveraging expert-based approaches.

First, \citet{wang2024human} examined annotator characteristics, where researchers compared individual annotator judgments against expert-provided labels rather than against other annotators, arguing that consensus does not always equate to correctness. Their findings showed that high inter-annotator agreement can sometimes mask low-quality or superficial judgments, particularly when annotators share common biases or lack subject matter expertise. Researchers further introduced a predictive modeling approach to identify reliable annotators in advance, using background traits (e.g., education, domain familiarity) and behavioral features (e.g., response time). By combining expert-aligned accuracy with annotator profiling, this work offers a novel lens for establishing annotation reliability independent of peer agreement. 

Second, \citet{nahum2024llms} established a high-confidence ground truth and rigorously addressed IRR challenges. They implemented an expert-based re-annotation protocol in conjunction with LLM-driven error detection. They used 11 datasets on a number of binary tasks, such as identifying (or not) hallucinations in LLM-generated text and fact verification.  Specifically, researchers focused expert efforts on examples where LLM ensemble predictions disagreed with the original labels, regardless of model confidence. Each of these examples was independently reviewed by two expert annotators—who were familiar with task definitions and annotation guidelines. During the initial annotation phase, both experts rated examples on a binary scale, classifying them as either factually consistent or inconsistent. To ensure impartiality, examples were presented in randomized order without exposing the original or LLM-provided labels. Following this phase, a reconciliation process was undertaken for all examples where the annotators disagreed. The experts discussed each disagreement to reach consensus, allowing for a refined label set used throughout the remainder of our analysis. This reconciliation improved the quality of annotations and significantly enhanced IRR, with Fleiss’s $\kappa$ for expert annotations increasing from 0.486 to 0.851 after reconciliation.

For comparison, researchers computed IRR statistics across all annotator types, including crowd-sourced workers, individual LLMs (with prompt variations), and an ensemble of LLM models. The results, summarized in Table \ref{tab:annotation-kappa}, reveal that while GPT-4 and PaLM2 achieved high $\kappa$ values (0.706 and 0.750, respectively), closely matching human expert reliability, crowd-sourced annotations from MTurk exhibited near-random agreement ($\kappa$ = 0.074). Furthermore, the ensemble approach of combining multiple LLM models and prompts yielded a moderate $\kappa$ of 0.521, suggesting that assembling not only improves the label quality, but also stabilizes the level of agreement across annotations. These findings highlight the value of expert reconciliation in overcoming IRR limitations and underscore the relative strengths of LLMs as scalable annotation tools when expert supervision is applied selectively.

\begin{table}[t]
\centering
\small
\caption{Reliability across annotation sources demonstrating excellent agreement among experts after reconciliation. Originally published in \citet{nahum2024llms}.}
\renewcommand{\arraystretch}{1.1}
\begin{tabular}{l c l}
\hline
\textbf{Source} & \textbf{Fleiss’s $\kappa$} & \textbf{Interpretation} \\
\hline
Experts (Post-Recon.) & 0.851 & Excellent \\
GPT-4                 & 0.706 & Close to experts \\
PaLM2                 & 0.750 & High reliability \\
MTurk Workers         & 0.074 & Near random \\
\hline
\end{tabular}
\label{tab:annotation-kappa}
\end{table}

\subsection{Predictive Validity}
\textit{Reported Use Case: Mapping open responses to MCQs on the same learning objectives}. \citet{thomas2025improving} used predictive validity as an alternative method to IRR in validating the effectiveness of their large language models (LLM) on assessing tutor learners open responses while engaging in scenario-based training. In this context, to establish predictive validity, the researchers use multiple-choice questions (MCQs) that assess the same learning objectives as open-response questions \cite{thomas2025improving}. Figure 1 displays an open response and corresponding multiple-choice question assessing the same learning objective of how to effectively respond to a student who has just made a math error.

\begin{figure}
\includegraphics[width=0.48
\textwidth]{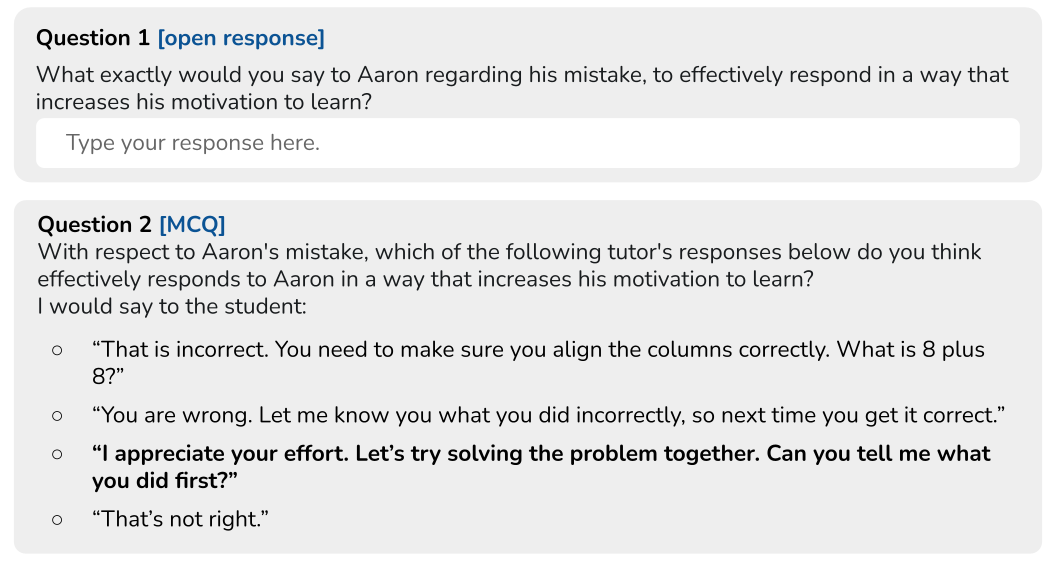}
\caption{ An open response and corresponding MCQ (correct selection in bold) that assess a tutor learner on the same learning objective (effectively responding to a student who has made a math error). By using predictive validity, human grading is not the sole source of ground truth. Adapted and modified from \citet{thomas2025improving,thomas2025does}.}
\end{figure}

Predictive validity refers to the extent to which an assessment accurately forecasts or correlates with future performance on a related measure \cite{trochim2016research}. In this case, predictive validity evaluates whether performance on MCQs can reliably predict outcomes on open-response questions or other measures of learner understanding. Much work has found MCQs can be as effective, and more efficient, than open response tasks when instructional time is limited \cite{thomas2025does,butler2018multiple}. This approach enhances objectivity by reducing subjective biases inherent in human scoring while ensuring that MCQs serve as effective proxies for more complex assessments. 
 
To assess the predictive validity of LLM-generated scores, namely GPT-4o, Gemini 1.5 pro, and LearnLM, on open responses in relation to MCQ scores, \citet{thomas2025improving} computed the correlation of participants’ MCQ scores and their LLM scores on open responses. The analysis revealed a significant, positive correlation between MCQ scores and human-graded open-response scores, \(r(86) = 0.421, p < .001\). While this correlation was statistically significant, it was not particularly large. Potential contributors to the moderate correlation size can be attributed to the few test items and inherent ambiguity in judging the correctness of open responses, even for human graders. Given they were striving to find alternative methods to human grading, this is not the best direct comparison. Thus, they computed the correlation between MCQ and LLM scores, correlating MCQ scores with GPT-4o-scored open responses, yielding \(r(86) = 0.406, p < .001 \); and MCQ scores with LearnLM-scored open responses, yielding \( r(86) = 0.477, p < .001 \). They found that LLM scores have significant predictive validity and this validity can be determined without open response grading by humans.

Notably, the LearnLM correlation of 0.477 is 0.056 higher than the human-scored correlation of 0.421. In other words, with this tighter comparison, we find that the predictive validity of the LLM scoring is comparable to that of human scoring. This result supports predictive validity as a complementary method for evaluating model performance, though in this particular case, it should be combined with additional measures to ensure a more comprehensive assessment. Other alternative approaches to using human scores as “ground truth” mentioned that align with LLMs-as-a judge include: using the average LLM score among several models, including adversarial models, e.g., LearnLM vs GPT-4o, to establish reliability, rather than comparing human judgments; and applying LLM self-consistency measures by comparing evaluations across varying prompts to check robustness.

\subsection{Close-the-loop Validity}
\textit{Reported Use Case: Mapping tutor move classifications to student performance and learning outcomes}. From a measurement point of view, where we have critiqued the use of human-derived annotation schemes and codes as limited, close-the-loop validity can also be used to qualify the predictive capabilities of a measurement or label. Specifically, close-the-loop validity ensures that a given assessment or model produces improved learning in line with its theoretical underpinning or coding scheme. A recent example of this is \citet{wang2024tutor} who demonstrated that tutors with access to Tutor CoPilot—an AI-powered system trained to reflect specific expert pedagogical strategies and reasoning—were more likely to use strategies aligned with high-quality teaching (e.g., asking guiding questions), and that these differences in tutor behavior translated into significant gains in student mastery, particularly for students taught by lower-rated tutors. This illustrates how the measurement of pedagogical quality and its operationalization in tutor strategy recommendations, supported by AI-based classifiers, can exhibit close-the-loop (and internal) validity by linking to the learning outcomes of the students within the system. Crucially, by linking tutor practices to student learning gains internally, in addition to validating classifiers, the researchers closed the loop on their classification scheme and demonstrated that their evidence-based taxonomy of tutor strategy recommendations correlates meaningfully with better learning. However, the study was also not without limitations: in particular, they did not correlate the specific occurrence or frequency of strategies (e.g., identified in tutor transcripts) to differences in learning gains. As we go on, such correlations of individual labels (and their dosage) with learning may pose an even stronger form of close-the-loop validity.

\section{Towards a Multidimensional Ground Truth and Future Directions}

Establishing internal validity among AI classifiers, such as the case of Tutor CoPilot, is uncommon, and desperately needed in the field of educational AI. But what is even more uncommon is external validity. \textit{\textbf{We could not find a single example use case demonstrating external validity within educational AI.}} Establishing external validity is rare in educational AI. External validity is a type of validity, which broadly refers to how well a measurement or study captures what it intends to measure and supports the conclusions drawn from it \cite{trochim2016research}. Specifically, external validity is about generalizability—whether findings from one context apply to other people, settings, or times. In the case of tutoring, external validity may ask whether assessments of tutoring skills during upskilling (such as in structured training tasks or simulations) accurately reflect how tutors will perform in real-world practice. It also concerns whether observed effectiveness in one context (e.g., producing an effective response during training) can be expected in others. Without external validity, we cannot assume that skills shown in training or outcomes observed in one setting will transfer to different, authentic tutoring environments.

Part of the reason for the lack of external validity is that real-world evaluation studies are costly and resource-intensive. It requires the time of teachers, often school permissions, and other overhead that purely algorithmic evaluations of assessment models do not reach. This is perhaps also why they are so important. While we can make use of evidence-based practices to grade tutor moves and create taxonomies that we know will likely make a difference in practice (e.g., prioritizing self-explanation in tutoring, which is known to enhance learning in lab experiments \cite{berthold2009assisting}), such practices are not guaranteed to lead to improved learning in authentic classroom contexts. We argue that these forms of external validity do matter and can reveal gaps between technical innovation and real-world impact in education. We believe that generalizability matters. 

In 2019, Baker presented a list of six challenges within the field of learning analytics \cite{baker2019challenges}. The description of these challenges included the evidence needed to demonstrate that the challenge was solved. In a similar fashion, we propose establishing generalizability of different AI tutoring classifiers across datasets as a challenge.

The details of the challenge are described as follows: 

1) Build AI classifiers to identify or detect tutor moves;

2) Apply these classifiers across tutoring datasets of diverse tutor-student populations and varying tutoring modalities and implementations;

3) Provide evidence by demonstrating that the classifiers work across datasets, e.g., with degradation of quality under 0.1 (AUC ROC, Pearson/Spearman correlation, and remaining better than chance). 

We leave this challenge to researchers and developers of tutoring models within educational AI. %

\section{Summary and Conclusion}

Automated assessment methods and AI require increasingly large amounts of human-annotated training data in education. As these AI systems increasingly shape how learning is assessed and supported, the validity of their training data becomes ever more critical. Though tempting for straightforward validation and quantifiable metrics of fidelity, relying solely on IRR to define “ground truth” risks reinforcing flawed human judgments rather than optimizing for meaningful educational outcomes.

This paper challenges the field’s prominent reliance on IRR as the primary standard for validating annotations in educational AI. We argue that this reliance often overlooks the complexity, subjectivity, and pedagogical significance of human responses, especially in open-ended or dialogic tasks. By showcasing supplemental frameworks, such as multi-label annotation, expert-based reconciliation, predictive validity, and close-the-loop experimentation—we demonstrate that richer, more reliable forms of “ground truth” are possible. 

Moving forward, educational AI should prioritize multidimensional and validity-centered approaches, striving for external validity, to ensure its tools are not only scalable but also meaningful, effective, and grounded in authentic learning outcomes.

\bibliography{custom}

\end{document}